\documentclass[sigconf,natbib=false]{acmart}
\AtBeginDocument{%
  }

\setcopyright{acmlicensed}
\copyrightyear{2018}
\acmYear{2018}
\acmDOI{XXXXXXX.XXXXXXX}
\acmConference[Conference acronym 'XX]{Make sure to enter the correct
  conference title from your rights confirmation email}{June 03--05,
  2018}{Woodstock, NY}
\acmISBN{978-1-4503-XXXX-X/2018/06}

\RequirePackage[
  datamodel=acmdatamodel,
  style=acmnumeric,
  sorting=none,
  ]{biblatex}

\begin{document}

\title{Beyond Decision Boundaries: Relational Geometry Attacks on Contrastive Embedding Manifolds}

\author{Fei Zhao}
\affiliation{%
  \institution{The University of Alabama at Birmingham}
  \city{Birmingham}
  % \state{Alabama}
  \country{USA}
}
\email{larry5@uab.edu}

\author{Peiyuan Zhang}
\affiliation{%
  \institution{The University of Alabama at Birmingham}
  \city{Birmingham}
  % \state{Alabama}
  \country{USA}
}
\email{zhangp@uab.edu}

\author{Xi Li}
\affiliation{%
  \institution{The University of Alabama at Birmingham}
  \city{Birmingham}
  % \state{Alabama}
  \country{USA}
}
\email{xiliuab@uab.edu}

\author{Chengcui Zhang}
\affiliation{%
  \institution{The University of Alabama at Birmingham}
  \city{Birmingham}
  % \state{Alabama}
  \country{USA}
}
\email{czhang02@uab.edu}

\author{Nitesh Saxena}
\affiliation{%
  \institution{Texas A\&M University}
  \city{College Station}
  % \state{Texas}
  \country{USA}
}
\email{nsaxena@tamu.edu}

\begin{abstract}

Contrastive learning and Siamese embedding models have become the foundation of modern verification systems, where decisions are governed not by discrete classification boundaries, but by relational geometry in embedding space. Consequently, the security of such systems depends on the stability of pairwise similarity organization rather than conventional decision regions. However, existing adversarial attacks remain fundamentally classification-centric. They primarily manipulate isolated prediction outputs through boundary perturbations while largely overlooking the vulnerability of relational geometry itself. In this paper, we introduce a geometry-aware adversarial attack framework that reformulates attacks on contrastive systems as manifold-level relational corruption. Instead of targeting individual predictions, the proposed framework systematically distorts similarity organization within the embedding manifold by pushing positive pairs apart while simultaneously pulling negative pairs closer, ultimately collapsing and inverting pairwise similarity structure. To enable scalable deployment, we shift iterative online optimization into an offline adversarial geometry deformation prior learning stage and train a lightweight feed-forward generator that learns generalized geometry deformation patterns from the victim model. Once trained, the generator produces adversarial perturbations through a single forward pass without requiring online gradient computation, enabling real-time online attacks against similarity-based verification systems. Experimental results across multiple verification architectures demonstrate substantial degradation of verification performance together with severe manifold-level relational corruption. On the Markmatch verification system, the proposed attack reduces accuracy from 95.4\% to 38.6\% while completely reversing the positive-negative similarity structure. Our findings reveal that contrastive embedding systems possess fundamentally different adversarial vulnerabilities from traditional classification models, suggesting that future robustness in contrastive representation learning systems may depend less on protecting decision boundaries and more on preserving relational geometry itself.

\end{abstract}



% \keywords{Do, Not, Use, This, Code, Put, the, Correct, Terms, for,
  % Your, Paper}


\received{20 February 2007}
\received[revised]{12 March 2009}
\received[accepted]{5 June 2009}

\maketitle

\section{Introduction}

\begin{figure}[htbp]
    \centering
    \includegraphics[width=0.45\textwidth]{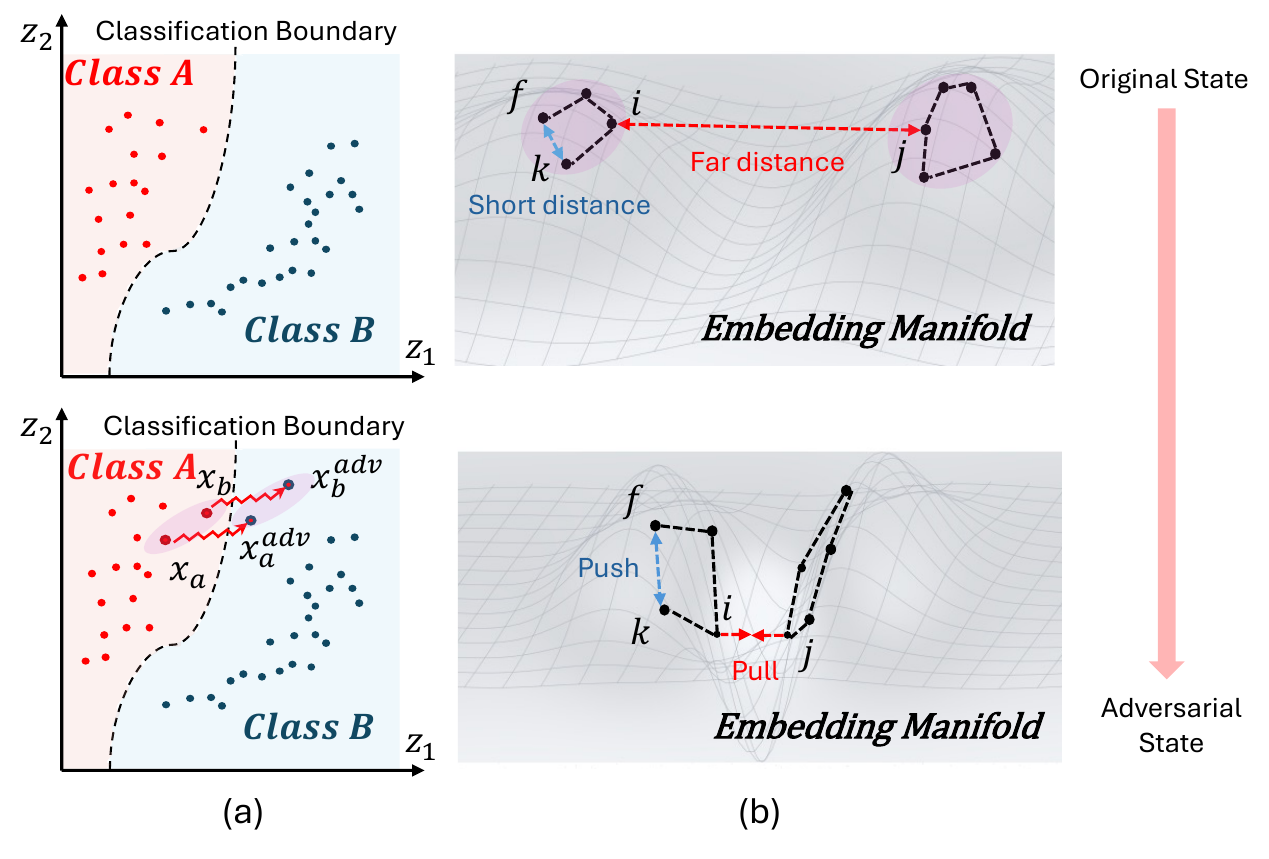}
    \caption{Conventional Attack (a): Traditional boundary-oriented perturbations fail to disrupt the continuous relational structure, leaving the relative pairwise distances (indicated by the persistent purple region) entirely intact. (b) Relational Geometry Attack (b): Our proposed framework directly warps the contrastive embedding manifold, successfully shattering the clean pairwise relationship to systematically invert similarity orderings (Push/Pull).}
    \label{fig:manifold}
\end{figure}

Contrastive learning \cite{chen2020simple,chopra2005learning} and Siamese networks \cite{bromley1993signature, koch2015siamese} have become the foundation of modern verification and retrieval systems, including biometric authentication \cite{wang2024ecg}, signature verification \cite{viana2022contrastive}, handwriting analysis \cite{li2026enhancing}, and ballot mark inspection \cite{zhao2024bubblesig}. Unlike traditional classification models that rely on discrete decision boundaries, these systems operate through relational geometry in embedding space \cite{zhao2025markmatch}. Inputs are projected into structured latent manifolds where verification decisions are governed entirely by pairwise similarity organization. Semantically similar samples are pulled together, while dissimilar samples are pushed apart. Consequently, the reliability of contrastive systems depends not on the stability of isolated decision regions for classes, but on the integrity of relational organization within the embedding manifold itself.

This paradigm shift in representation learning introduces a fundamentally different adversarial vulnerability. Traditional adversarial attacks are largely designed for classification systems, where the primary objective is to cross local classification decision boundaries and manipulate isolated prediction outputs, shown as Fig. \ref{fig:manifold}(a). However, contrastive systems do not fundamentally rely on discrete boundaries. Their behavior emerges from relative similarity structure distributed across the embedding space. As a result, attacks against contrastive models should not merely focus on label flipping or local prediction manipulation. Instead, they must target the relational geometry governing pairwise organization in latent space. The central vulnerability of contrastive systems therefore lies in the manipulability of embedding geometry itself.

Despite recent progress in adversarial machine learning, existing attack frameworks remain predominantly classification-centric. Iterative optimization methods such as Projected Gradient Descent (PGD) \cite{madry2017towards} repeatedly compute sample-specific perturbations through expensive online gradient optimization. While effective for classification-oriented objectives, such attacks exhibit two major limitations when applied to modern similarity systems. First, they fail to systematically corrupt global relational organization within embedding manifolds, shown as Fig. \ref{fig:manifold}(a). Second, their heavy online optimization overhead makes them impractical for real-time online verification environments requiring low-latency deployment. 

% Modern similarity-based infrastructures increasingly operate under streaming and large-scale conditions, where iterative optimization for every individual sample pair becomes computationally prohibitive.

In this paper, we introduce a geometry-aware adversarial attack framework that reformulates attacks on contrastive systems as manifold-level relational corruption. \textbf{Rather than targeting isolated prediction outputs, the proposed framework systematically distorts pairwise similarity organization by pushing positive pairs apart while simultaneously pulling negative pairs closer, ultimately collapsing and inverting relational structure within the embedding manifold}, shown as Fig. \ref{fig:manifold}(b). To enable scalable deployment, we shift expensive iterative online optimization into an offline adversarial geometry deformation prior learning stage. During this stage, a lightweight feed-forward generator learns generalized geometry deformation patterns from the victim similarity model. Once trained, the generator produces adversarial perturbations through a single forward pass without requiring online gradient computation, enabling real-time online attacks against similarity-based verification systems.

We evaluate the proposed framework across multiple security-critical contrastive verification architectures, including same-hand ballot mark verification \cite{zhao2025markmatch} and Siamese signature verification systems \cite{chokshi2023sigscatnet,dey2017signet}. Experimental results demonstrate substantial degradation of verification performance and systematic collapse of positive-negative separation in embedding space. Our findings suggest that contrastive embedding systems possess fundamentally different adversarial vulnerabilities from traditional classification models, highlighting the emerging security risks of manipulable relational geometry in modern representation learning systems.

\noindent\textbf{Our contributions are summarized as follows.}

\textbf{First}, we introduce a geometry-centric adversarial attack paradigm for contrastive systems that shifts the attack objective from local decision boundary crossing to manifold-level relational corruption in embedding space.

\textbf{Second}, we propose a lightweight generator-based framework that transforms expensive iterative online optimization into offline adversarial geometry deformation prior learning, enabling instantaneous attack generation through a single feed-forward pass without online gradient computation.

\textbf{Third}, we demonstrate that modern contrastive verification systems possess fundamentally different adversarial vulnerabilities from traditional classification models, where attacks can systematically collapse and invert pairwise similarity organization across embedding manifolds.

\section{Related Works}

Traditional adversarial attacks on deep neural networks are predominantly built upon classification-oriented learning paradigms. Foundational methods such as Fast Gradient Sign Method (FGSM) \cite{goodfellow2014explaining} and Projected Gradient Descent (PGD) \cite{madry2017towards} established the vulnerability of neural architectures by generating perturbations that push samples across discrete decision boundaries. Subsequent robustness benchmarks and evaluations \cite{wang2023better, croce2020reliable} further reinforced this boundary-centric adversarial perspective. These attacks are highly effective in closed-set recognition settings \cite{scheirer2012toward}, where model behavior is primarily governed by isolated prediction regions. However, their underlying objective remains fundamentally tied to manipulating individual prediction outputs through local classification boundary perturbations. As modern contrastive representation learning increasingly shifts toward similarity-driven systems, this classification-centric perspective becomes insufficient for characterizing the vulnerabilities of contrastive embedding models.

Recent advances in contrastive learning and metric learning have exposed emerging security challenges in similarity-based systems. Prior studies have investigated adversarial vulnerabilities in metric learning frameworks \cite{mao2019metric} and Siamese verification networks \cite{jahangir2023adversarial}. Existing approaches \cite{zhou2020adversarial, zaid2021ranking} typically adapt iterative optimization strategies to manipulate ranking objectives \cite{li2022arra} or disrupt triplet relationships \cite{jiang2024anti}. These works demonstrate that contrastive embedding spaces can indeed be adversarially compromised. Nevertheless, most existing attacks fundamentally inherit the optimization philosophy of classification-based adversarial learning. They primarily operate through sample-specific perturbation optimization and local similarity manipulation rather than systematically targeting the global relational organization of embedding space. Consequently, they fail to explicitly model adversarial corruption as a manifold-level geometry distortion problem, see Fig. \ref{fig:manifold}(b). In addition, their reliance on repeated backward passes and continuous model queries introduces substantial online optimization overhead, making them impractical for large-scale retrieval and real-time verification environments.

Generative adversarial attack frameworks provide an alternative direction for reducing the computational cost of iterative online optimization. Methods such as AdvGAN \cite{xiao2018generating} train neural generators to synthesize adversarial perturbations directly through feed-forward inference. Recent generative attack frameworks \cite{zhu2024ge, poursaeed2018generative, baluja2017adversarial} further improve attack efficiency and scalability in high-throughput settings by shifting expensive optimization into an offline learning stage. However, existing generative attacks remain largely constrained to conventional classification objectives \cite{sharma2022deep}. Their learned perturbation priors are primarily designed to induce classification label flipping, see Fig. \ref{fig:manifold}(a) rather than corrupt relational geometry in embedding space, see Fig. \ref{fig:manifold}(b). In contrast, our work reformulates adversarial attacks on contrastive systems as manifold-level relational corruption. Rather than learning perturbation priors for isolated prediction manipulation, the proposed geometry deformation generative network learns generalized geometry deformation patterns that systematically collapse and invert pairwise similarity organization across the embedding manifold while maintaining real-time online deployment efficiency.

\section{Methods}
\subsection{Problem: Relational Geometry Attack}

Contrastive verification systems fundamentally operate under a different paradigm from traditional classification models. Instead of producing independent class predictions, they organize semantic relationships through relational geometry in embedding space, where verification decisions are governed by the topology of the latent manifold rather than explicit decision boundaries.

Within the embedding manifold, semantically similar samples form coherent local neighborhoods, while dissimilar samples remain topologically separated. Consequently, the reliability of contrastive systems depends on preserving the global relational organization of the manifold itself. Under this perspective, adversarial attacks against contrastive systems should not merely manipulate isolated similarity scores. Instead, they should directly deform the manifold topology governing pairwise semantic organization. For positive pairs, the attack objective is to separate embeddings occupying nearby semantic regions. For negative pairs, the objective is to collapse originally separated manifold regions and distort their relational structure. The goal is therefore not class boundary crossing, but systematic corruption of pairwise similarity organization and relational geometry within the embedding space.

We define this attack paradigm as a \textit{Relational Geometry Attack}. Unlike conventional adversarial attacks that manipulate isolated prediction outputs, \textbf{a relational geometry attack seeks to systematically collapse and invert pairwise similarity organization in latent space by directly deforming the underlying embedding manifold}.

\subsection{Manifold Deformation Framework}

Given an input pair $(A,B)$ with pair label $y$, the proposed framework learns adversarial geometry deformation priors that systematically deform the relational geometry of the embedding manifold while preserving the overall visual appearance of the inputs. Rather than optimizing sample-specific perturbations through iterative online optimization, the framework learns generalized geometry deformation patterns capable of corrupting pairwise semantic organization across the latent space.

The adversarially perturbed samples are forwarded through the frozen victim embedding model, where verification is performed through pairwise similarity relationships in embedding space. The attack objective is formulated as manifold-level relational corruption. For positive pairs, the framework seeks to separate embeddings occupying nearby semantic regions. For negative pairs, it seeks to collapse originally separated manifold regions and distort their relational boundaries. The resulting deformation systematically disrupts pairwise similarity organization and inverts relational structure across the embedding manifold, shown as Fig. \ref{fig:manifold}(b).

\subsection{Threat Model}

\textbf{Attacker's abilities.} The attack consists of offline training and an online attack.
\textbf{Offline training} follows the classic adversarial optimization setting~\cite{madry2017towards}.
The attacker independently collects a small auxiliary dataset $\mathcal{D}_{\mathrm{attack}}^{\mathrm{train}}$ that follows a similar distribution to, but has no overlap with, the victim model's training data $\mathcal{D}_{\mathrm{victim}}^{\mathrm{train}}$, i.e., 
\[
\mathcal{D}_{\mathrm{victim}}^{\mathrm{train}} \cap \mathcal{D}_{\mathrm{attack}}^{\mathrm{train}} = \varnothing,
\]
and uses it solely to train the attack generator.
The attacker also has white-box access to the frozen victim model, enabling extraction of embedding representations, pairwise similarity logits, and gradient signals via backpropagation.
However, the attacker does not modify the victim model's parameters, architecture, or training data throughout the attack generator training process.
The \textbf{online attack} operates under a black-box setting.
The attacker independently collects query data $\mathcal{D}_{\mathrm{attack}}^{\mathrm{test}}$ and generates adversarial example pairs via the trained attack generator, which are then submitted to the victim model through queries without receiving any internal model information.

\noindent\textbf{Attacker's goals.}
The attacker aims to invert the victim model's pairwise similarity judgments: positive pairs should be classified as dissimilar, and negative pairs as similar.
The adversarial perturbations added to the image pairs are imperceptible to human observers, making the attack difficult to detect.

\subsection{Geometry Deformation Generative Network}
\label{subsec:GenerativeNetwork}

To operationalize relational geometry attacks, we introduce a noval geometry-aware deformation generative network that learns generalized manifold deformation patterns against contrastive verification systems. Rather than performing expensive sample-specific optimization during inference, the proposed framework amortizes manifold deformation into a feed-forward generative process capable of systematically distorting relational organization within the embedding space.

Formally, let $G_{\theta}$ denote the geometry deformation generator parameterized by $\theta$. Given an input pair $(A,B)$, the generator produces corresponding geometry-aware perturbation fields for both inputs:
\[
    \tilde{\delta}_A = G_{\theta}(A),
    \quad
    \tilde{\delta}_B = G_{\theta}(B).
\]
These learned perturbations are added to the original inputs to produce adversarially perturbed samples whose embeddings systematically corrupt relational organization within the latent manifold. Although the perturbations are generated at the image level, the optimization objective is defined entirely over pairwise semantic organization in embedding space. The generator therefore learns generalized relational corruption priors that systematically separate semantically aligned regions while collapsing originally separated pairwise structures within the embedding manifold.

The deformation generator is implemented using a U-Net-based encoder-decoder architecture with skip connections \cite{ronneberger2015unet}. U-Net is adopted due to its ability to model structured spatial transformations while preserving fine-grained local information. Unlike conventional generative adversarial attacks that primarily learn perturbation priors for classification label manipulation \cite{xiao2018generating,poursaeed2018generative,baluja2017adversarial}, the proposed generator is explicitly optimized to deform manifold topology and corrupt relational geometry within contrastive embedding space. To keep the input-space perturbations bounded, we constrain the generated geometry-aware perturbation fields using both element-wise clipping and $L_2$ projection:
\[
    \delta_A =
    \Pi_{\epsilon_2}
    \left(
    \mathrm{clip}_{[-\epsilon_{\infty}, \epsilon_{\infty}]}
    (\tilde{\delta}_A)
    \right),
    \quad
    \delta_B =
    \Pi_{\epsilon_2}
    \left(
    \mathrm{clip}_{[-\epsilon_{\infty}, \epsilon_{\infty}]}
    (\tilde{\delta}_B)
    \right),
\]
where $\epsilon_{\infty}$ controls the maximum pixel-level perturbation and $\epsilon_2$ constrains the overall $L_2$ perturbation budget. The adversarial inputs are then constructed as
\[
    A^{adv} =
    \mathrm{clip}_{[0,1]}(A + \delta_A),
    \quad
    B^{adv} =
    \mathrm{clip}_{[0,1]}(B + \delta_B).
\]
The clipping operation ensures valid image formation while preserving visually subtle perturbations. Consequently, the proposed framework learns bounded geometry deformation patterns that remain visually constrained while inducing substantial relational corruption within the embedding manifold.

\subsection{Victim Model}
\label{subsec:victim}
Pairwise verification models are commonly implemented with Siamese-style architectures, where two inputs are processed by shared or tied feature extractors and compared in a learned representation space \cite{chopra2005learning,bromley1993signature,koch2015siamese}. 
The victim model in this work is a frozen pairwise verification model in the MarkMatch same-hand ballot mark verification framework \cite{zhao2025markmatch}. It measures the similarity of an input sample pair in the embedding space. Given an input pair $(A, B)$, the victim model uses a shared image encoder to extract the embedding representations of the two images. Its encoder consists of a DenseNet121 \cite{huang2017densely} backbone followed by a linear embedding layer. The original classification head of DenseNet121 is removed, and global average pooling is used to extract image features. The resulting feature vector is then mapped to a 64-dimensional embedding space by a dense layer. The parameters of the victim model remain frozen during the training of the geometry deformation generative network.

Formally, let $f_{\phi}$ denote the victim encoder, where $\phi$ represents the parameters of the victim model. For a clean input pair $(A, B)$, the embeddings are given by:
\[
    z_A = f_{\phi}(A),
    \quad
    z_B = f_{\phi}(B).
\]
To compute pairwise similarity, the two embeddings are first $L_2$-normalized:
\[
    \hat{z}_A =
    \frac{z_A}{\|z_A\|_2},
    \quad
    \hat{z}_B =
    \frac{z_B}{\|z_B\|_2}.
\]
The victim model then computes cosine similarity as the dot product between the normalized embeddings:
\[
    s(A,B) =
    \hat{z}_A^\top \hat{z}_B.
\]
Following standard contrastive learning formulations \cite{radford2021learning}, the cosine similarity is further scaled by a temperature parameter $\tau$ to obtain the similarity logit:
% In implementation, the similarity score is scaled by a temperature parameter $\tau$ to obtain the similarity logit:
\[
    \ell(A,B) =
    \frac{s(A,B)}{\tau}.
\]
Since $\tau < 1$, the resulting logits amplify relative similarity differences while preserving the original similarity ordering in embedding space. Therefore, all reported logits in the experiments can be interpreted as temperature-scaled cosine similarities. For an adversarial pair $(A^{adv}, B^{adv})$, the victim model computes the adversarial similarity logit in the same way:
\[
    \ell(A^{adv}, B^{adv})
    =
    \frac{1}{\tau}
    \left(
    \frac{f_{\phi}(A^{adv})}{\|f_{\phi}(A^{adv})\|_2}
    \right)^\top
    \left(
    \frac{f_{\phi}(B^{adv})}{\|f_{\phi}(B^{adv})\|_2}
    \right).
\]
In evaluation, the similarity logit is further converted into a probability using the sigmoid function and compared with a predefined threshold to obtain the pairwise verification decision. During the training of the geometry deformation generative network, the victim model provides embeddings, similarity logits, and gradient signals for backpropagation. The parameters of the victim model remain frozen, and only the parameters of the geometry deformation generative network are updated.

\subsection{Attack Optimization Objective}

The attack objective is designed to induce relational corruption within the embedding manifold of the frozen victim model. The proposed framework directly manipulates pairwise semantic organization by distorting similarity relationships in latent space. For each input pair $(A,B)$ with pair label $y$, we first compute the clean similarity logit and the adversarial similarity logit:
\[
    \ell_{\mathrm{clean}} = \ell(A,B),
    \quad
    \ell_{\mathrm{adv}} = \ell(A^{adv}, B^{adv}).
\]
The relational displacement induced by the deformation generator is then defined as
\[
    \Delta \ell =
    \ell_{\mathrm{adv}} - \ell_{\mathrm{clean}}.
\]
This displacement serves as a differentiable surrogate for relational geometry corruption in embedding space. Since verification decisions in Siamese-style systems are governed by pairwise similarity organization rather than independent class predictions \cite{bromley1993signature,koch2015siamese}, the desired direction of relational displacement depends on the pair label. For positive pairs $(y=1)$, the objective is to reduce similarity and separate embeddings occupying nearby semantic regions. For negative pairs $(y=0)$, the objective is to increase similarity and collapse originally separated manifold regions. To encode this label-dependent relational deformation direction, we define
\[
    \alpha(y) =
    \begin{cases}
        1, & y = 1, \\
        -1, & y = 0.
    \end{cases}
\]
The main relational corruption objective is formulated as
\[
    \mathcal{L}_{\mathrm{attack}}
    =
    \frac{1}{N}
    \sum_{i=1}^{N}
    w_{y_i}\alpha(y_i)\Delta \ell_i,
\]
where $N$ is the batch size and $w_{y_i}$ denotes the class-dependent weighting factor. Minimizing this objective encourages relational separation for positive pairs and relational collapse for negative pairs, thereby systematically reducing the semantic separation structure of the embedding manifold.

To further stabilize manifold-level deformation behavior across a batch, we introduce a relational direction regularization term. Let $\overline{\Delta \ell}_{+}$ denote the average relational displacement over positive pairs and let $\overline{\Delta \ell}_{-}$ denote the average displacement over negative pairs. The regularization term is defined as
\[
    \mathcal{L}_{\mathrm{dir}}
    =
    \lambda_{\mathrm{dir}}
    \left[
    \mathrm{ReLU}(\overline{\Delta \ell}_{+})
    +
    \mathrm{ReLU}(-\overline{\Delta \ell}_{-})
    \right].
\]
where $\lambda_{\mathrm{dir}}$ is a hyperparameter controlling the strength of the relational direction regularization term. This term penalizes inconsistent manifold deformation directions when positive pairs fail to separate or negative pairs fail to collapse.

We additionally include an optional perturbation regularization term based on the $L_2$ norm of the generated geometry-aware perturbation fields:
\[
    \mathcal{L}_{\mathrm{pert}}
    =
    \frac{1}{2}
    \left(
    \frac{1}{N}\sum_{i=1}^{N}\|\delta_{A,i}\|_2
    +
    \frac{1}{N}\sum_{i=1}^{N}\|\delta_{B,i}\|_2
    \right).
\]
The final optimization objective is
\[
    \mathcal{L}
    =
    \mathcal{L}_{\mathrm{attack}}
    +
    \mathcal{L}_{\mathrm{dir}}
    +
    \lambda_{\mathrm{pert}}\mathcal{L}_{\mathrm{pert}}.
\]
During optimization, the victim model remains fixed while the geometry deformation generative network learns relational corruption patterns over the latent similarity structure. In practice, $\lambda_{\mathrm{pert}}$ can be set to zero since perturbation magnitude is already constrained through clipping and projection operations, introduced in Section \ref{subsec:victim}.

\section{Experiments}

We evaluate the proposed framework across multiple verification architectures and analyze both verification performance degradation and geometric corruption of latent similarity organization.

\subsection{Datasets and Victim Models}

% dataset
We conduct experiments on two datasets.
The ballot mark pair dataset from MarkMatch~\cite{zhao2025markmatch} consists of $51 \times 51$ three-channel image pairs, where each sample is an image pair $(A, B)$ with a binary pair label $y \in \{0, 1\}$: $y=1$ indicates a positive pair (two ballot marks considered similar or from the same hand) and $y=0$ a negative pair (two ballot marks considered dissimilar or from different hands).
The dataset is balanced, with approximately equal numbers of positive and negative pairs.
The CEDAR~\cite{kalera2004offline} signature verification dataset contains genuine and forged handwritten signature pairs, similarly represented as $(A, B)$ with a binary pair label, and is widely used for pairwise verification benchmarking.
For both datasets, we follow the default training and test split, yielding $\mathcal{D}_{\mathrm{victim}}^{\mathrm{train}}$ and $\mathcal{D}_{\mathrm{victim}}^{\mathrm{test}}$ respectively.
For the ballot mark pair dataset, the training set $\mathcal{D}_{\mathrm{victim}}^{\mathrm{train}}$ contains 16,272 pairs and the test set $\mathcal{D}_{\mathrm{victim}}^{\mathrm{test}}$ contains 1,972 pairs.

The main victim model follows the MarkMatch framework~\cite{zhao2025markmatch}, built on a DenseNet121 encoder~\cite{huang2017densely} pretrained on ImageNet. The encoder maps each $51 \times 51$ RGB image to a 64-dimensional $L_2$-normalized embedding, and the pairwise logit is computed as the cosine similarity scaled by temperature $\tau = 0.07$.
The model is trained with a CLIP-style bidirectional batch contrastive loss~\cite{radford2021learning} combined with a binary cross-entropy loss on positive pairs, with equal weights on both terms.
Training uses the Adam optimizer~\cite{kingma2014adam} with a learning rate of $10^{-4}$, batch size of 32, and 10 epochs. The verification threshold is selected on the validation ROC curve via Youden's $J$ statistic~\cite{youden1950index}, yielding $\sim 0.9962$.
Besides, we also evaluate against contrastive learning-based verification systems, including SigNet~\cite{dey2017signet}, SigScatNet~\cite{chokshi2023sigscatnet}, and a CEDAR DenseNet verification model~\cite{cedar_signature}.

\subsection{Attack Setup}

Following the threat model, we randomly split $\mathcal{D}_{\mathrm{victim}}^{\mathrm{test}}$ into the auxiliary dataset $\mathcal{D}_{\mathrm{attack}}^{\mathrm{train}}$ for attack generator offline training and $\mathcal{D}_{\mathrm{attack}}^{\mathrm{test}}$ for online attack evaluation at an 8:2 ratio, ensuring no overlap with the victim model's training and test data:
\[
    \mathcal{D}_{\mathrm{victim}}^{\mathrm{test}}
    =
    \mathcal{D}_{\mathrm{attack}}^{\mathrm{train}}
    \cup
    \mathcal{D}_{\mathrm{attack}}^{\mathrm{test}},
    \quad
    \mathcal{D}_{\mathrm{attack}}^{\mathrm{train}}
    \cap
    \mathcal{D}_{\mathrm{attack}}^{\mathrm{test}}
    =
    \varnothing .
\]
For the ballot mark pair dataset, $\mathcal{D}_{\mathrm{victim}}^{\mathrm{test}}$ contains 1,972 pairs, yielding $\mathcal{D}_{\mathrm{attack}}^{\mathrm{train}}$ with 1,578 pairs and $\mathcal{D}_{\mathrm{attack}}^{\mathrm{test}}$ with 394 pairs.

The attack generator, introduced in Section~\ref{subsec:GenerativeNetwork}, is a U-Net-based feed-forward model~\cite{ronneberger2015unet} that takes a single image as input and outputs a perturbed version. For each input pair $(A, B)$, the generator is \textbf{applied independently to each image}, yielding the adversarial pair $(A^{\mathrm{adv}}, B^{\mathrm{adv}})$.
During offline training, the generator is trained on $\mathcal{D}_{\mathrm{attack}}^{\mathrm{train}}$ for 400 epochs with a batch size of 256 and a learning rate of $10^{-4}$. The perturbation is constrained by an $L_2$ budget of $\epsilon_2 = 2.5$, which bounds the overall perturbation magnitude, and an element-wise clipping threshold of $\epsilon_{\mathrm{pix}} = 16/255$, which limits per-pixel intensity changes. Together, these constraints ensure the adversarial perturbations remain imperceptible to human observers, following standard adversarial attack settings~\cite{goodfellow2014explaining, madry2017towards}. The positive and negative pair weights in the attack objective are both set to 1.0.
During the online attack, adversarial pairs are generated from $\mathcal{D}_{\mathrm{attack}}^{\mathrm{test}}$ following the same perturbation procedure described above, and submitted to the victim model as queries.

\subsection{Baselines and Evaluation Metrics}

We compare the proposed relational geometry attacks with three baseline methods: (1) Diff-PGD \cite{xue2023diffusion}, an iterative gradient-based attack adapted to the pairwise verification setting. It extends the PGD-style adversarial optimization paradigm \cite{madry2017towards} with diffusion-guided adversarial sample generation. (2) A limited-query black-box attack based on zeroth-order optimization with NES-style gradient estimation, denoted as ZO/NES \cite{ilyas2018blackbox}. This baseline does not directly use victim-model gradients. Instead, it estimates attack directions through repeated queries to the victim model. Zeroth-order and query-limited black-box attacks estimate adversarial directions from model outputs without directly accessing model gradients \cite{chen2017zoo}. (3) Surrogate Transfer. In this setting, adversarial examples are generated using a substitute verification model and then evaluated on the victim model. This baseline is motivated by substitute-model-based black-box transfer attacks, where adversarial examples crafted on a local substitute model can transfer to the target model \cite{papernot2017practical}.

All methods are evaluated using the same attack test split. In the main evaluation, we report the Clean-Clean setting and the Adv-Adv setting. Clean-Clean evaluates the victim model on clean input pairs, while Adv-Adv evaluates the victim model when both images in each pair are attacked. \textbf{The attack is applied independently to all images rather than selectively targeting specific pairs or images, so that every pairwise comparison is performed using adversarially perturbed inputs.} Therefore, We do not include Clean-Adv or Adv-Clean in the experiments.

% We do not include Clean-Adv or Adv-Clean in the main tables because our goal is to evaluate pair-level adversarial corruption when the full input pair is attacked.

We use verification accuracy as the decision-level metric. Accuracy is computed by converting the similarity logit into a probability and comparing it with the fixed verification threshold. Since accuracy alone does not fully describe how the pairwise similarity structure changes, we also report three relational-geometry metrics.

Regarding positive pairs, we compute Positive Logit Drop:
\[
    \mathrm{PosDrop}
    =
    \ell_{\mathrm{clean}}^{+}
    -
    \ell_{\mathrm{adv}}^{+},
\]
where $\ell_{\mathrm{clean}}^{+}$ and $\ell_{\mathrm{adv}}^{+}$ denote the average positive-pair logits before and after the attack. A larger Positive Logit Drop means that originally similar pairs are pushed farther apart.

Regarding negative pairs, we compute Negative Logit Rise:
\[
    \mathrm{NegRise}
    =
    \ell_{\mathrm{adv}}^{-}
    -
    \ell_{\mathrm{clean}}^{-},
\]
where $\ell_{\mathrm{clean}}^{-}$ and $\ell_{\mathrm{adv}}^{-}$ denote the average negative-pair logits before and after the attack. A larger Negative Logit Rise means that originally dissimilar pairs are pulled closer together.

Finally, we compute Gap Reduction:
\[
    \mathrm{GapReduction}
    =
    \left(
    \ell_{\mathrm{clean}}^{+}
    -
    \ell_{\mathrm{clean}}^{-}
    \right)
    -
    \left(
    \ell_{\mathrm{adv}}^{+}
    -
    \ell_{\mathrm{adv}}^{-}
    \right).
\]
This metric measures how much the attack reduces the separation between positive and negative pairs. A larger Gap Reduction indicates stronger corruption of the relational geometry in the embedding space.

\section{Results and Ablation}

This section evaluates whether the proposed framework can systematically corrupt relational geometry in contrastive verification systems. The experiments are designed to examine manifold-level semantic separation collapse, pairwise similarity inversion, and relational structure deformation within the embedding space. We first analyze the proposed framework on the primary MarkMatch victim model and compare its relational corruption behavior against existing adversarial baselines. We then investigate whether similar relational geometry vulnerabilities emerge across different contrastive verification architectures. Finally, we study the effects of bounded deformation constraints, transformation-based defenses, and amortized attack generation efficiency.

\subsection{Main Results and Baseline Comparison}
As shown in Table~\ref{tab:main_detailed}, the proposed framework produces substantial relational corruption on the primary DenseNet121 victim model. Under clean inputs, the victim model achieves an accuracy of 0.954, with an average positive-pair logit of 9.993 and an average negative-pair logit of 0.641. This large positive-negative separation indicates that the embedding manifold preserves a strong semantic relational structure prior to attack. After adversarial deformation, the verification accuracy decreases to 0.386, accompanied by a severe collapse of relational separation within the embedding space. The average positive-pair logit decreases from 9.993 to -2.478, while the average negative-pair logit increases from 0.641 to 3.340, where \textbf{the logits correspond to temperature-scaled pairwise similarity scores} defined in Section~\ref{subsec:victim}. Consequently, the original semantic similarity organization becomes inverted: \textbf{embeddings that originally occupied nearby manifold regions are pushed apart, while originally separated manifold regions collapse toward each other.} This result demonstrates that the proposed framework does not merely reduce verification accuracy, but fundamentally corrupts the relational geometry governing the embedding manifold.

\begin{table}[H]
\centering
\caption{Main attack performance}
\label{tab:main_detailed}
\resizebox{\columnwidth}{!}{
\begin{tabular}{lrrrrr}
\toprule
Scenario & Acc $\downarrow$ & Pos Logit $\downarrow$ & Neg Logit $\uparrow$ & Pos Prob $\downarrow$ & Neg Prob $\uparrow$ \\
\midrule
Clean & 0.954 & 9.993 & 0.641 & 0.996 & 0.559 \\
Adv & 0.386 & -2.478 & 3.340 & 0.326 & 0.589 \\
\bottomrule
\end{tabular}
}
\end{table}

To further quantify manifold-level relational corruption, we analyze the relational separation statistics summarized in Table~\ref{tab:main_derived}. The proposed framework produces a Positive Logit Drop of 12.471 and a Negative Logit Rise of 2.698, indicating simultaneous separation of semantically aligned manifold regions and collapse of originally separated relational structures. More importantly, the clean positive-negative relational gap decreases from 9.352 to -5.818 after attack, resulting in a Gap Reduction of 15.170. Since the adversarial relational gap becomes negative, the original semantic separation structure is completely inverted: \textbf{negative pairs become more similar than positive pairs within the embedding manifold}. This result demonstrates that the proposed framework does not merely perturb pairwise similarity scores locally, but systematically collapses and reverses the global relational organization governing the latent embedding space.

\begin{table}[H]
\small
\centering
\caption{Relational geometry metrics}
\label{tab:main_derived}
\begin{tabular}{lr}
\toprule
Metric & Value \\
\midrule
Accuracy Drop & 0.569 \\
Positive Logit Drop & 12.471 \\
Negative Logit Rise & 2.698 \\
Clean Gap & 9.352 \\
Adv Gap & -5.818 \\
Gap Reduction & 15.170 \\
\bottomrule
\end{tabular}
\end{table}

As shown in Table~\ref{tab:baseline_summary}, the proposed framework is compared against existing adversarial baselines. Although several baselines partially perturb pairwise similarity relationships, the proposed framework produces substantially stronger manifold-level relational corruption across all evaluated metrics. Diff-PGD reduces the adversarial accuracy to 0.660 and achieves a Gap Reduction of 5.458, indicating partial degradation of semantic separation structure. ZO/NES exhibits only limited relational corruption, with an adversarial accuracy of 0.919 and a Gap Reduction of 1.200. Although Surrogate Transfer produces stronger relational displacement than ZO/NES, its relational collapse remains substantially weaker than the proposed framework, achieving a Gap Reduction of 4.584. \textbf{In contrast,} the proposed framework reduces adversarial accuracy to 0.386 while producing a Gap Reduction of 15.170, substantially exceeding all baseline methods. More importantly, only the proposed framework fully reverses the original positive-negative relational organization within the embedding manifold. These results suggest that existing attacks primarily induce local similarity perturbations, whereas the proposed geometry-aware framework systematically collapses semantic separation structure at the manifold level.

\begin{table}[H]
\centering
\caption{Comparison with baselines}
\label{tab:baseline_summary}
\resizebox{\columnwidth}{!}{
\begin{tabular}{lrrrrrr}
\toprule
Method & Clean Acc & Adv Acc $\downarrow$ & Acc Drop $\uparrow$ & Pos Drop $\uparrow$ & Neg Rise $\uparrow$ & Gap Red. $\uparrow$\\
\midrule
Diff-PGD & 0.954 & 0.660 & 0.294 & 4.842 & 0.616 & 5.458 \\
ZO/NES & 0.954 & 0.919 & 0.035 & 0.091 & 1.109 & 1.200 \\
Surrogate Transfer & 0.954 & 0.736 & 0.218 & 3.756 & 0.829 & 4.584 \\
\textbf{Ours} & {0.954} & \textbf{0.386} & \textbf{0.569} & \textbf{12.471} & \textbf{2.698} & \textbf{15.170} \\
\bottomrule
\end{tabular}
}
\end{table}

Table~\ref{tab:baseline_detailed} provides a more detailed view of how different attacks affect relational organization within the embedding manifold. Since all methods share the same clean reference on the MarkMatch victim model, the clean statistics are reported only once. Diff-PGD produces partial relational displacement by decreasing the positive-pair logit from 9.993 to 5.152 and increasing the negative-pair logit from 0.641 to 1.258. However, the positive-pair similarity remains substantially higher than the negative-pair similarity, indicating that the original semantic separation structure is weakened but still preserved. ZO/NES primarily increases the negative-pair logit while leaving the positive-pair structure nearly unchanged, suggesting limited local similarity perturbation without substantial manifold-level relational restructuring. Surrogate Transfer perturbs both positive and negative directions, but its relational deformation remains insufficient to collapse the original semantic organization. In contrast, the proposed framework produces a complete relational inversion on the DenseNet121 victim model. The positive-pair logit decreases to -2.478, while the negative-pair logit increases to 3.340, resulting in negative pairs becoming more similar than positive pairs within the embedding manifold. Among all evaluated attacks, only the proposed geometry-aware framework fully collapses and reverses the original semantic separation structure governing the latent relational organization.

\begin{table}[H]
\centering
\caption{Adversarial statistics by method}
\label{tab:baseline_detailed}
\resizebox{\columnwidth}{!}{
\begin{tabular}{lrrrrr}
\toprule
Method & Adv Acc $\downarrow$ & Pos Logit $\downarrow$ & Neg Logit $\uparrow$ & Pos Prob $\downarrow$ & Neg Prob $\uparrow$ \\
\midrule
Clean Reference & 0.954 & 9.993 & 0.641 & 0.996 & 0.559 \\
\midrule
Diff-PGD & 0.660 & 5.152 & 1.258 & 0.835 & 0.592 \\
ZO/NES & 0.919 & 9.902 & 1.750 & 0.997 & 0.678 \\
Surrogate Transfer & 0.736 & 6.238 & 1.470 & 0.959 & 0.670 \\
\textbf{Ours} & \textbf{0.386} & \textbf{-2.478} & \textbf{3.340} & \textbf{0.326} & \textbf{0.589} \\

\bottomrule
\end{tabular}
}
\end{table}

\subsection{Cross System Evaluation}

We further evaluate whether manifold-level relational corruption emerges consistently across different contrastive verification architectures. As shown in Table~\ref{tab:cross_system}, the proposed framework substantially degrades relational organization across all evaluated systems, although the degree of manifold deformation varies across architectures and datasets. On SigNet \cite{dey2017signet}, the clean verification accuracy decreases from 1.000 to 0.500 under adversarial deformation, indicating severe corruption of the underlying similarity structure. SigScatNet \cite{chokshi2023sigscatnet} exhibits an even stronger relational collapse, where the adversarial accuracy decreases from 0.982 to 0.324 and the Gap Reduction reaches 17.760, demonstrating substantial destruction of semantic separation structure within the embedding manifold. The CEDAR DenseNet experiment follows a similar pattern, with adversarial accuracy decreasing from 0.970 to 0.501 and a Gap Reduction of 10.254.

Despite architectural differences across victim models, all evaluated systems exhibit consistent relational geometry degradation under the proposed framework. \textbf{These results suggest that the observed vulnerability does not arise from a specific backbone architecture alone, but instead reflects an intrinsic weakness of similarity-driven representation learning systems }whose decisions depend on preserving manifold-level relational organization.

\begin{table}[H]
\centering
\caption{Cross system evaluation}
\label{tab:cross_system}
\resizebox{\columnwidth}{!}{
\begin{tabular}{lrrrr}
\toprule
Model & Clean Acc & Adv Acc $\downarrow$ & Acc Drop $\uparrow$ & Gap Red. $\uparrow$\\
\midrule
SigNet & 1.000 & 0.500 & 0.500 & 8.478 \\
SigScatNet & 0.982 & 0.324 & 0.658 & 17.760 \\
CEDAR DenseNet & 0.970 & 0.501 & 0.470 & 10.254 \\
\bottomrule
\end{tabular}
}
\end{table}

\subsection{Perturbation Budget Ablation}

We analyze the role of the $L_2$ perturbation budget while fixing the pixel level clipping threshold at $\epsilon_{\mathrm{pix}}=16/255$. The main configuration uses $\epsilon_2=2.5$. To understand the effect of the $L_2$ constraint, we also evaluate a setting with a very large $L_2$ budget, which approximates a case without an effective $L_2$ constraint. In this setting, the perturbation is mainly controlled by the pixel level clipping threshold. Both configurations remain effective. With $\epsilon_2=2.5$, the adversarial accuracy is 0.386 and the Gap Reduction is 15.170. Without an effective $L_2$ constraint, the adversarial accuracy is 0.378 and the Gap Reduction is 14.720. The similar performance indicates that the pixel level clipping threshold already provides a strong constraint, while the $L_2$ projection offers additional control on the overall perturbation magnitude.

The consistently large Gap Reduction across both settings suggests that manifold-level relational collapse does not depend on large unconstrained perturbations. Instead, substantial semantic separation corruption can still emerge under tightly bounded perturbations. These results indicate that the observed relational vulnerability is fundamentally tied to the embedding-space organization of contrastive systems rather than merely the magnitude of input-space perturbations.

\begin{table}[H]
\small
\centering
\caption{$L_2$ budget ablation}
\label{tab:budget_ablation}
\begin{tabular}{lrrrr}
\toprule
$L_2$ Budget & Adv Acc & Pos Drop $\uparrow$ & Neg Rise $\uparrow$ & Gap Red. $\uparrow$ \\
\midrule
$\epsilon_2=2.5$ & 0.386 & 12.471 & 2.698 & 15.170 \\
No $L_2$ constraint & 0.378 & 12.427 & 2.293 & 14.720 \\
\bottomrule
\end{tabular}
\end{table}

\begin{figure}[H]
\centering
\includegraphics[width=0.7\columnwidth]{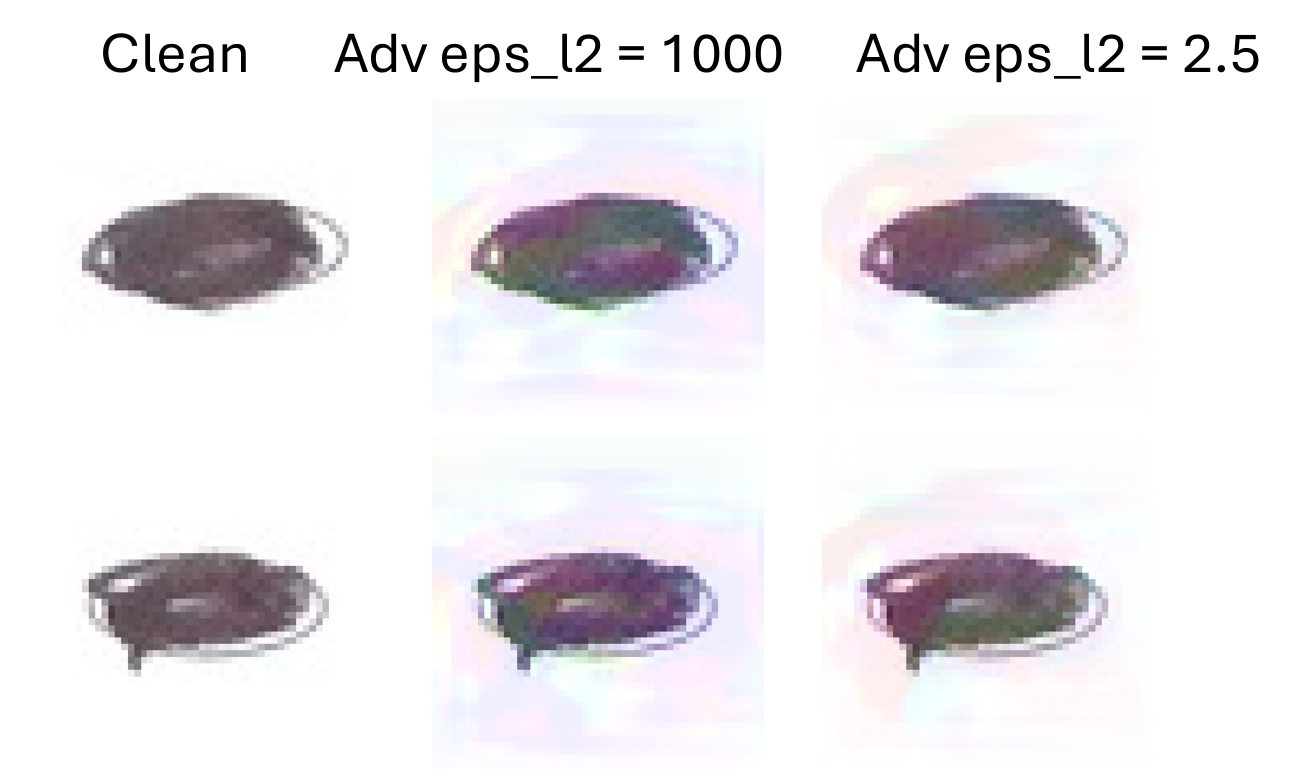}
\caption{Qualitative comparison of clean and adversarial samples under different $L_2$ perturbation budgets. A very large $L_2$ budget produces more visible color artifacts, while the main setting with $\epsilon_2=2.5$ better preserves the visual appearance of the clean ballot marks.}
\label{fig:l2_visualization}
\end{figure}

Fig.~\ref{fig:l2_visualization} provides a qualitative comparison between clean images and adversarial images generated under different $L_2$ budgets. When the $L_2$ budget is set to a very large value, the perturbations become more visually noticeable and introduce stronger color artifacts around the ballot marks. In contrast, the main setting with $\epsilon_2=2.5$ produces adversarial samples that better preserve the visual appearance of the clean inputs while still maintaining strong attack effectiveness. This supports the use of the constrained $L_2$ budget in the main experiments.

\subsection{Transformation Defense and Query Cost}

We further evaluate whether a simple input-space transformation based defense can mitigate manifold-level relational corruption induced by the proposed framework. In this setting, random input transformations are applied before verification. This preprocessing strategy can be viewed as a lightweight defense because it changes the input distribution and may partially disrupt the generated perturbations.

As shown in Table~\ref{tab:transform_robustness}, random transformations weaken the attack but do not eliminate relational geometry corruption.  Without transformation, the adversarial accuracy decreases to 0.363, with an accuracy drop of 0.591 and a Gap Reduction of 16.550. With random transformations, the adversarial accuracy increases to 0.480, and the Gap Reduction decreases to 10.094. Although the defense reduces the severity of relational collapse, the embedding manifold still exhibits substantial semantic separation degradation under adversarial deformation.

\begin{table}[H]
\centering
\caption{Transformation based defense}
\label{tab:transform_robustness}
\resizebox{\columnwidth}{!}{
\begin{tabular}{lrrrrrr}
\toprule
Setting & Clean Acc & Adv Acc $\downarrow$ & Acc Drop $\uparrow$ & Pos Drop $\uparrow$ & Neg Rise $\uparrow$ & Gap Red. $\uparrow$\\
\midrule
No Transform & 0.954 & 0.363 & 0.591 & 13.529 & 3.021 & 16.550 \\
Random Transform & 0.939 & 0.480 & 0.459 & 8.038 & 2.056 & 10.094 \\
\bottomrule
\end{tabular}
}
\end{table}

The proposed framework generates perturbations through a single feed-forward pass and therefore avoids repeated victim-model optimization during inference. In contrast, the ZO/NES black-box baseline estimates attack directions through iterative online queries to the victim verification model, requiring repeated exploration of the embedding-space similarity landscape. In the limited-query black-box setting, attacking 1,972 pairs requires 851,904 attack-phase queries, corresponding to 428 optimization queries per pair. Including pre-attack verification and final evaluation queries, the complete attack pipeline requires 861,764 total victim-model queries and a runtime of 20,682 seconds, as summarized in Table~\ref{tab:query_cost}. These results highlight the substantial computational overhead of iterative online relational attacks against contrastive embedding systems. In contrast, the proposed framework amortizes manifold deformation into an offline geometry-learning stage, enabling real-time relational corruption without repeated embedding-space optimization during deployment.

\begin{table}[H]
\small
\centering
\caption{ZO/NES query cost}
\label{tab:query_cost}
\begin{tabular}{lr}
\toprule
Metric & Value \\
\midrule
Pairs attacked & 1972 \\
Runtime & 20682 s \\
Pre-attack verification queries & 3944 \\
Attack-phase queries & 851904 \\
Final evaluation queries & 5916 \\
Total queries & 861764 \\
Optimization queries per pair & 428 \\
Overall attack success & 0.656 \\
\bottomrule
\end{tabular}
\end{table}

\section{Conclusion}

In this paper, we introduced a geometry-aware adversarial framework for contrastive verification systems and Siamese embedding models. Unlike conventional adversarial attacks that primarily target discrete classification boundaries, the proposed framework reformulates adversarial manipulation as manifold-level relational corruption in embedding space. Rather than attacking isolated prediction outputs, the proposed approach systematically deforms relational geometry by separating semantically aligned manifold regions while collapsing originally separated similarity structures.

By shifting iterative optimization into an offline manifold deformation learning stage, the proposed generator learns generalized geometry corruption patterns that enable real-time relational attacks through a single forward pass. Experimental results across multiple verification architectures demonstrate substantial degradation of pairwise verification performance, severe corruption of relational similarity organization, and systematic collapse of semantic separation structure within the embedding manifold.

Our findings suggest that the adversarial vulnerability of contrastive systems fundamentally differs from that of traditional classification models. As modern AI systems increasingly rely on similarity-driven representation learning, retrieval, and embedding-based reasoning, future robustness may depend less on protecting decision boundaries and more on preserving the relational geometry governing pairwise similarity organization itself.

% \begin{acks}
% This work was supported in part by NSF CNS-2154589, 2154443, and 2154507, “Collaborative Research: SaTC: CORE: Medium: Bubble Aid: Assistive AI to Improve the Robustness and Security of Reading Hand-Marked Ballots,” \$1,200,000, 10/01/2022-09/30/2026.
% \end{acks}

% \bibliographystyle{ACM-Reference-Format}
% \bibliography{ref}
\printbibliography

\end{document}